\documentclass[10pt]{article} 
\usepackage[accepted]{tmlr}

\usepackage{amsmath,amsfonts,bm}

\def\eqref#1{equation~\ref{#1}}

\def\1{\bm{1}}

\DeclareMathAlphabet{\mathsfit}{\encodingdefault}{\sfdefault}{m}{sl}
\SetMathAlphabet{\mathsfit}{bold}{\encodingdefault}{\sfdefault}{bx}{n}

\usepackage{hyperref}
\usepackage{url}
\usepackage[utf8]{inputenc} 
\usepackage[T1]{fontenc}    
\usepackage{hyperref}       
\usepackage{url}            
\usepackage{booktabs}       
\usepackage{amsfonts}       
\usepackage{nicefrac}       
\usepackage{microtype}      
\usepackage{xcolor}         

\usepackage{adjustbox}
\usepackage{multirow}
\usepackage{colortbl}
\usepackage{amsmath}
\usepackage{tabularx}
\usepackage{subcaption}

\usepackage{graphicx}
\usepackage{graphics}   
\usepackage{amssymb}
\usepackage{balance}
\usepackage{adjustbox}
\usepackage{algorithm}
\usepackage{algorithmic}
\usepackage{tikzducks}
\usepackage{pifont}

\newcommand{\Fig}[1]{Figure~\ref{fig:#1}}

\newcommand{\Tbl}[1]{Table~\ref{tab:#1}}

\def\ie{\textit{i.e.}}

\definecolor{brown}{rgb}{0.85, 0.15, 0.15}
\definecolor{purp}{rgb}{0.95, 0.16, 0.65}
\definecolor{purpc}{rgb}{0.95, 0.36, 0.65}
\definecolor{orange}{rgb}{0.9, 0.45, 0.0}
\definecolor{blue}{rgb}{0.0, 0.5, 1.0}
\definecolor{green}{rgb}{0, 0.8, 0}
\definecolor{lgreen}{rgb}{0.6, 0.8, 0}
\definecolor{red}{rgb}{0.8, 0, 0}
\definecolor{redd}{rgb}{0.9, 0, 0}
\definecolor{yellow}{rgb}{0.75, 0.56, 0}
\definecolor{darkblue}{rgb}{0.2, 0.2, 0.8}
\definecolor{brinkpink}{rgb}{0.98, 0.38, 0.5}
\definecolor{cadmiumred}{rgb}{0.89, 0.0, 0.13}
\definecolor{ceruleanblue}{rgb}{0.16, 0.32, 0.75}
\definecolor{dandelion}{rgb}{0.94, 0.88, 0.19}
\definecolor{bostonuniversityred}{rgb}{0.8, 0.0, 0.0}
\definecolor{brown(web)}{rgb}{0.65, 0.16, 0.16}
\definecolor{cornellred}{rgb}{0.7, 0.11, 0.11}
\definecolor{greend}{rgb}{0.0, 0.35, 0.0}

\newcommand{\sohyun}[1]{{\color{black}{#1}}}

\definecolor{lblue}{rgb}{0, 0.2, 0.8}
\definecolor{dorange}{rgb}{0.8, 0.4, 0.0}

\def\etal{\textit{et al.}}

\newcommand{\bx}{\mathbf{x}}

\usepackage{array}
\newcolumntype{C}[1]{>{\centering\arraybackslash}p{#1}}

\newcommand{\hyperfootnote}[1][]{\def\ArgI\hyperfootnoteRelay}
\newcommand\hyperfootnoteRelay[2][]{\href{#1#2}{\ArgI}\footnote{\href{#1#2}{#2}}}

\usepackage{enumitem}

\title{TestDG: Test-time Domain Generalization \\
for Continual Test-time Adaptation}

\author{\name Sohyun Lee \email lshig96@postech.ac.kr \\
      \addr POSTECH
      \AND
      \name Nayeong Kim \email kimnay@postech.ac.kr \\
        \addr POSTECH
        \AND
        \name Juwon Kang \email juwon.kang@gengen.ai \\
        \addr GenGenAI 
        \AND
        \name Seong Joon Oh \email coallaoh@kaist.ac.kr \\
        \addr KAIST
        \AND
        \name Suha Kwak \email suha.kwak@postech.ac.kr \\
        \addr POSTECH
      }

\def\month{10}  
\def\year{2026} 
\def\openreview{\url{https://openreview.net/forum?id=qZ1HuY07Ud}} 

\begin{document}

\maketitle

\begin{abstract}
This paper studies continual test-time adaptation (CTTA), the task of adapting a model to constantly changing unseen domains in testing while preserving previously learned knowledge.
Existing CTTA methods mostly focus on adaptation to the current test domain only, overlooking generalization to arbitrary test domains a model may face in the future.
To tackle this limitation, we present a novel online test-time domain generalization framework for CTTA, dubbed TestDG.
TestDG aims to learn features invariant to both current and previous test domains on the fly during testing, improving the potential for effective generalization to future domains.
To this end, we propose a new model architecture and a test-time adaptation strategy dedicated to learning domain-invariant features, along with prototype selection and update mechanisms for effectively managing information from previous test domains.
TestDG achieved state of the art on three of four public CTTA benchmarks and performed on par with the best prior method on ImageNet-C.
Moreover, it showed superior generalization to unseen test domains.
\end{abstract}

\section{Introduction}

Deep neural networks have driven significant advances in various vision tasks such as classification~\citep{resnet,Li2019_SKNet,dosovitskiy2020image}, object detection~\citep{Rcnn, carion2020end, cai2018cascade}, and semantic segmentation~\citep{deconvnet,xie2021segformer,deeplab_v3}.
Despite these achievements, they often struggle with limited generalization ability to the domain shift between training and test data~\citep{ganin2015unsupervised,sun2016deep,chang2019domain}, such as image corruptions~\citep{Hendrycks2019_ImageNetC,chen2024robustsam,lee2025garasam}, adverse weather~\citep{Sakaridis_2018_IJCV,ma2022both,lee2022fifo,lee2024frest}, and low-light conditions~\citep{Sakaridis_2019_ICCV,lis,lee2023human}.
Test-time adaptation (TTA)~\citep{zhao2023pitfalls,Tent,zhang2022memo,niu2022efficient,niu2023towards} has been developed to mitigate the distribution shift by adapting a pre-trained model to unlabeled test domains during testing.
Specifically, TTA methods update the model in testing on the fly, by self-training using pseudo-labeled test data~\citep{song2023ecotta,gan2023decorate,liu2023vida} or by entropy minimization of the model prediction~\citep{niu2022efficient,Tent}.
Early approaches to TTA assume a single and fixed test domain. 
In reality, however, this assumption does not hold since a model is often deployed in non-stationary and continually changing environments.
In such environments, the accuracy of the model can easily degrade even with TTA since the constant distribution shift in testing causes unreliable pseudo labels~\citep{guo2017calibration}, which in turn exacerbates error accumulation and catastrophic forgetting.

\begin{figure}[t]
\centering
    \centering
    \includegraphics[width=\linewidth]{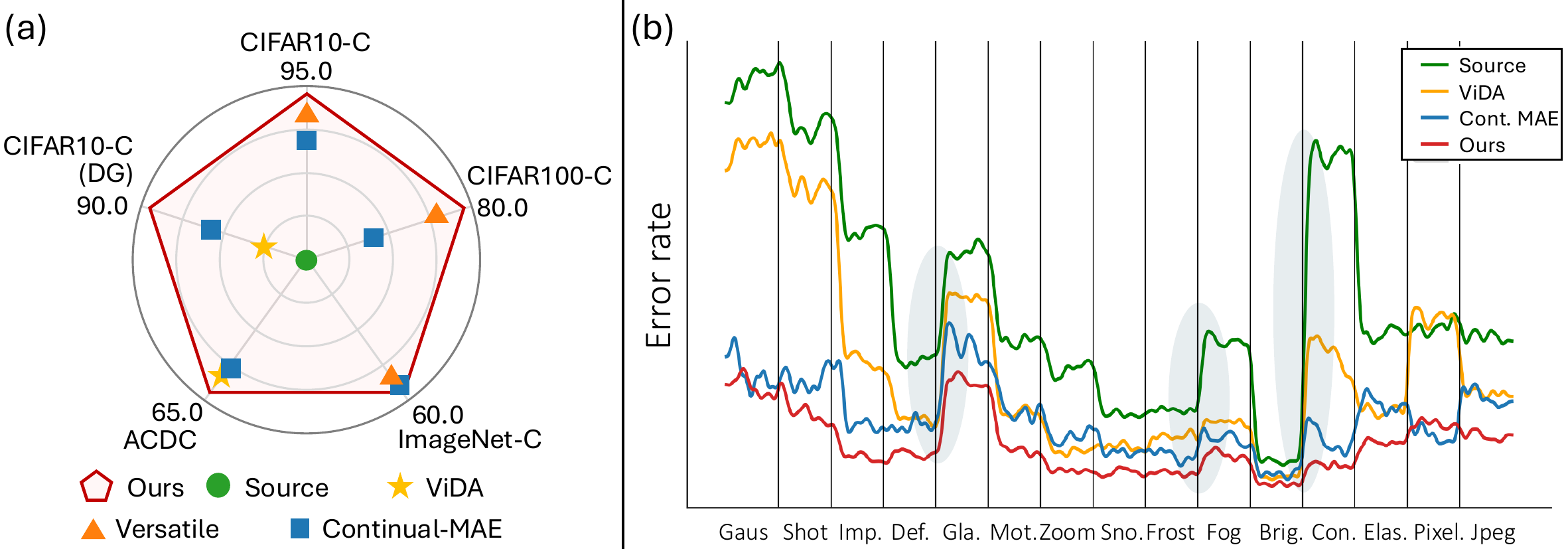}
    \caption{Empirical demonstration of the effectiveness of TestDG. (a) Accuracy across four CTTA and one domain generalization (DG) benchmarks, where DG performance is evaluated on 5 unseen domains after CTTA over 10 domains. For ACDC, mIoU is used instead. 
    (b) Error rate over sequentially changing test domains. 
    The consistently low error rates and notably stable performance of TestDG, especially in moments of domain change, highlight its robustness.}
    \label{fig:performance_teaser}
\end{figure}

Continual test-time adaptation (CTTA)~\citep{wang2022continual,song2023ecotta,gan2023decorate,liu2023vida,lee2024becotta,yang2023exploring,liu2024continual,yang2024versatile,zhu2025reshaping} has been studied to address these realistic TTA scenarios.
Most of existing CTTA methods focus only on adapting a model to the test domain at hand. Although these methods lead to notable performance improvements, there remains room for further improvement in that they do not fully account for the model's generalization to future test domains it encounters later in continuously changing environments.

In this paper, we present an online TEST-time Domain Generalization framework for continual test-time adaptation, dubbed TestDG.
TestDG overcomes the aforementioned limitation by learning domain-invariant representations that well generalize to unseen domains on the fly during testing.
This approach improves robustness of a model to arbitrary test domains that it may encounter in the future, as demonstrated in \Fig{performance_teaser}(a).
Furthermore, as shown in \Fig{performance_teaser}(b), learned domain-generalizable features enable robust adaptation even under abrupt domain shifts.
A na\"ive approach to implementing domain-invariant learning is to align features from different test domains.
However, since the features contain their own semantic contents as well as domain-specific information, this alignment without consideration of semantics corrupts the features and consequently degrades performance.

Hence, we introduce a \emph{domain information extractor} that takes the encoder features as input and extracts only domain-specific information in the form of embedding vectors; we call such vectors \emph{domain embeddings}.
The encoder is, in turn, trained to make domain embeddings from different domains indistinguishable by reducing the gap between such domain embeddings, so that it learns domain-invariant representation.
Analogous to adversarial learning~\citep{goodfellow2020generative,ganin2015unsupervised,li2018deep}, we alternate the optimization of the domain information extractor and that of the encoder. 
This approach allows the encoder to gradually become domain-invariant, and the domain information extractor to become sensitive to the remaining domain-specific information of the encoder.

The remaining challenge in implementing TestDG is that only the test domain at hand is accessible at each iteration of CTTA, while domain-invariant learning demands domain embeddings of at least two test domains.
To tackle this, TestDG efficiently stores embedding vectors that represent domain embeddings of the previous test domain; we call such embedding vectors \textit{domain prototypes} and sample them from domain embeddings of the previous domain through submodular optimization~\citep{nemhauser1978analysis,kim2016examples}.
By encouraging that domain embeddings of the current domain are not distinguished from the domain prototypes of the previous domain, TestDG performs domain-invariant learning of the encoder without access to samples of previous test domains.

TestDG was evaluated on the four public benchmarks for CTTA: CIFAR10-to-CIFAR10-C~\citep{cifar,Hendrycks2019_ImageNetC}, CIFAR100-to-CIFAR100-C~\citep{cifar,Hendrycks2019_ImageNetC}, ImageNet-to-ImageNet-C~\citep{Imagenet,Hendrycks2019_ImageNetC}, and Cityscapes-to-ACDC~\citep{cityscapes,acdc}.
It achieved a new state of the art on CIFAR10-C, CIFAR100-C, and Cityscapes-to-ACDC, performed on par with the best prior method on ImageNet-C, and demonstrated greater generalization ability on unseen test domains.

\section{Related work}
\noindent\textbf{Test-time adaptation}
Recent research on 
TTA
aims at addressing the domain shift problem by adapting a pre-trained model online during testing~\citep{zhao2023pitfalls,TTT,Tent,wang2022continual,liu2023vida,zhang2022memo}.
Initial studies~\citep{TTT,TTT++} employed self-supervised learning with proxy tasks on unlabeled test data, but 
they
require modifications to the training stage to accommodate the proxy tasks.
Meanwhile, Tent~\citep{Tent}, a fully test-time adaptation method based on entropy minimization, 
requires no modifications to the training stage and is applicable to any pre-trained model.
Building on this work, several studies~\citep{niu2022efficient, niu2023towards} have proposed using entropy-based losses in testing environments.
Other approaches~\citep{su2022revisiting,TTT++,eastwood2021source} leveraged lightweight information from source data to facilitate adaptation while mitigating catastrophic forgetting.
Unlike them, our method directly extracts domain embeddings from streamed test data, without computation for source data.

\noindent\textbf{Continual test-time adaptation}
In real-world scenarios, testing environments can vary significantly due to multiple factors such as weather, time, and geolocation.
Since conventional TTA methods have been studied under the assumption that the test domain is static, they face challenges in continually changing environments.
To address this limitation, \citet{wang2022continual} introduced the first CTTA method, leveraging multiple augmented inputs and a momentum network to generate more reliable pseudo labels.
\citet{song2023ecotta} and \citet{lee2024becotta} proposed memory-efficient methods that utilized lightweight meta networks and a mixture of experts, respectively.
\citet{gan2023decorate} and \citet{liu2023vida} utilized visual prompt learning and visual domain adapters, respectively, to disentangle domain-specific and domain-shared knowledge.
\citet{liu2024continual} improved target domain knowledge extraction through a self-supervised framework.
In this context, \citet{yang2024versatile} proposed a versatile network that identifies and refines unreliable pseudo-labels using the source pre-trained model, while \citet{zhu2025reshaping} introduced an uncertainty-aware buffering and graph-based constraint.
Unlike these methods, TestDG learns domain-invariant features that well generalize to diverse unseen test domains.

\noindent\textbf{Domain generalization}
Domain generalization (DG) focuses on improving the generalization performance of models on unseen target domains by training them with multiple source domains.
Initial work in DG employed domain alignment~\citep{li2018deep,shao2019multi,jia2020single,li2020domain,wang2021respecting} to learn domain-invariant representations by minimizing the distance of distribution among multiple domains.
Recent studies in DG have introduced various strategies, including data augmentation~\citep{zhou2020deep,zhou2020learning,kang2022style}, meta learning~\citep{balaji2018metareg,dou2019domain}, and representation learning~\citep{nam2021reducing,harary2022unsupervised,lee2022fifo}.
Inspired by DG, this paper presents a CTTA method that updates an online model through domain-invariant learning to handle continually changing domains.

\begin{figure*}[t]
\centering
\includegraphics[width=1.0\linewidth]{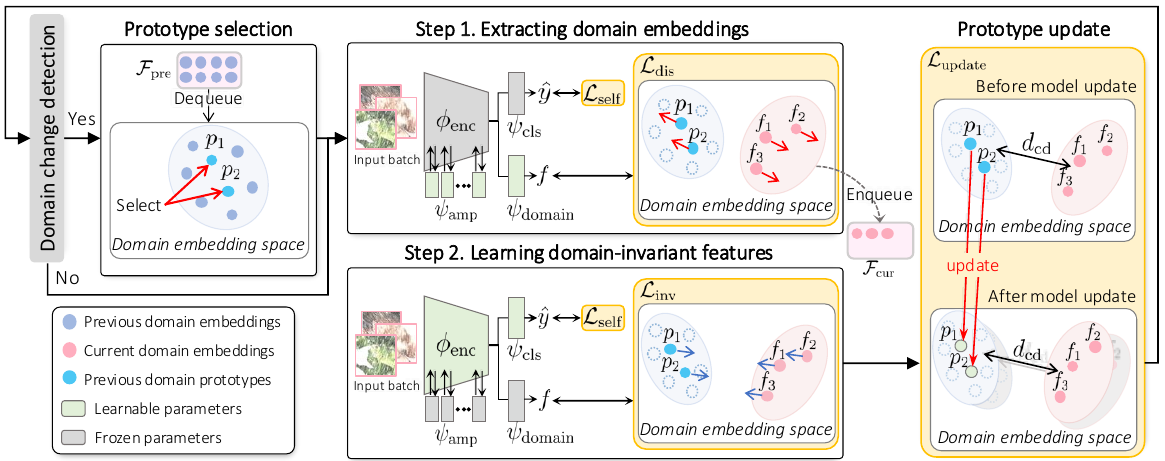}
\caption{
Overview
of TestDG.
When a domain change is detected, TestDG selects domain prototypes that well represent the previous domain embeddings stored in a queue. 
Each iteration consists of two steps.
In Step 1, the domain information extractor $\psi_\textrm{domain}$ learns alongside domain amplifier $\psi_\textrm{amp}$ to extract domain embeddings from the encoder features using the domain discrimination loss $\mathcal{L}_\textrm{dis}$.
In Step 2, the encoder $\phi_\textrm{enc}$ is trained using the domain-invariant loss $\mathcal{L}_\textrm{inv}$ to reduce the gap between domain embeddings and domain prototypes 
of the previous domain.
The domain prototypes are continuously updated to match with the updated model parameters by $\mathcal{L}_\textrm{update}$.
}
\label{fig:architecture}
\end{figure*}

\section{Proposed method}
TestDG is an online test-time domain generalization framework for CTTA, designed to learn domain-invariant features from a stream of test-time inputs;
its overall pipeline is illustrated in \Fig{architecture}.
Unlike domain adaptation and generalization, which in general assume that data from at least two domains are available during the entire training process, CTTA allows access to a handful of data from the test domain at hand (\ie, the input data at each test point) for each iteration.
This constraint poses a non-trivial challenge for domain-invariant learning.

TestDG addresses this challenge by storing \textit{domain embeddings}, \ie, embedding vectors containing domain information of the most recent previous domain, in a queue. 
To prevent excessive memory consumption caused by storing all domain embeddings, we select representative embeddings from the queue of domain embeddings as domain prototypes when the test domain changes.

To achieve domain-invariant learning without corrupting the semantics of features, TestDG operates in two steps as follows. 
In the first step, the domain information extractor learns to extract domain embeddings, by capturing only domain-specific information from the encoder features.
To facilitate this extraction, another extra module called \emph{domain amplifier} is attached to the encoder for capturing and emphasizing the domain-specific information of the encoder features.
We design the domain amplifier as a lightweight adapter structure~\citep{adaptformer,houlsby2019parameter,pfeiffer2020adapterfusion} to minimize additional parameters.
During this step, the domain amplifier is updated alongside the domain information extractor while the encoder remains frozen.
In the second step, we perform domain-invariant learning by ensuring the domain embeddings of the current domain become indistinguishable from the domain prototypes of the previous domain.
During this step, only the encoder is updated, with the domain information extractor and domain amplifier frozen.

By alternating between the two steps, the encoder gradually learns domain-invariant features across multiple test domains.
Meanwhile, the domain information extractor and domain amplifier are updated to reflect the update of the encoder and extract domain embeddings corresponding to the current state of the encoder.
In addition, TestDG continuously updates the domain prototypes to match them with the updated model parameters. 
Since the model used to extract domain prototypes is constantly updated during CTTA, the prototypes must also be updated accordingly.

\subsection{Domain change detection}\label{sec:domain_change}
As it is unknown when the test domain changes in CTTA, we detect domain changes based on prediction confidence scores as in \citet{gan2023decorate}.
That is, if the mean prediction confidence of the current batch differs from that of either of the two preceding batches by more than a threshold, the test domain is considered to have changed.
TestDG is designed to remain effective even with 
detection errors:
false positives are caused by intra-domain shifts and thus let TestDG learn invariant features within a single domain, while false negatives suggest mild distribution shifts that can be handled robustly by the model at hand as-is.

\subsection{Domain prototype selection}\label{sec:proto_selection}
When a domain change is detected, TestDG retains a small set of $n$ domain prototypes $\mathcal{P}_\textrm{pre}=\{p_i\}_{i=1}^{n}$ from the previous domain embeddings in the queue $\mathcal{F}_\textrm{pre}=\{f_i\}_{i=1}^{m}$ to reduce memory footprint substantially.
To ensure that these prototypes represent the entire distribution of the previous domain, 
we select $\mathcal{P}_\textrm{pre}$ by minimizing the discrepancy between $\mathcal{F}_\textrm{pre}$ and $\mathcal{P}_\textrm{pre}$.
Following the prototype sampling technique~\citep{kim2016examples}, we measure the discrepancy between $\mathcal{F}_\textrm{pre}$ and $\mathcal{P}_\textrm{pre}$ using the squared maximum mean discrepancy (MMD): 
\begin{align}    
\textrm{MMD}^2(\mathcal{F}_\textrm{pre},\mathcal{P}_\textrm{pre})  &\, := \frac{1}{|\mathcal{F}_\textrm{pre}|^2}\sum_{f_i,f_j \in \mathcal{F}_\textrm{pre}} k(f_i,f_j)  
     - \frac{2}{|\mathcal{F}_\textrm{pre}||\mathcal{P}_\textrm{pre}|}\sum_{f_i\in \mathcal{F}_\textrm{pre}, p_j\in \mathcal{P}_\textrm{pre}} k(f_i,p_j) \nonumber\\
    &\,    + \frac{1}{|\mathcal{P}_\textrm{pre}|^2}\sum_{p_i,p_j\in \mathcal{P}_\textrm{pre}} k(p_i,p_j) \;,
\end{align} 
where $k(\bx, \bx') =\text{exp}(-\gamma||\bx - \bx'||^2)$ is a RBF
kernel that measures the discrepancy between embeddings from the two sets.
As the first term in Eq.~(1) remains constant with respect to $\mathcal{P}_\textrm{pre}$, we define a score function $J(\mathcal{P}_\textrm{pre})$ as follows:
\begin{align}
    J(\mathcal{P}_\textrm{pre}) &\, := \text{MMD}^2(\mathcal{F}_\textrm{pre},\emptyset) - \text{MMD}^2(\mathcal{F}_\textrm{pre},\mathcal{P}_\textrm{pre}) \nonumber\\
        &\,   = \frac{2}{|\mathcal{F}_\textrm{pre}||\mathcal{P}_\textrm{pre}|}\sum_{f_i\in \mathcal{F}_\textrm{pre}, p_j \in \mathcal{P}_\textrm{pre}} k(f_i,p_j) - \frac{1}{|\mathcal{P}_\textrm{pre}|^2}\sum_{p_i,p_j \in \mathcal{P}_\textrm{pre}} k(p_i,p_j) \;,
\label{eq:cost}
\end{align} 
where the constant term $\text{MMD}^2(\mathcal{F}_\textrm{pre},\emptyset)$ is added to ensure that $J(\emptyset) = 0$.
The domain prototypes are selected by maximizing the score function:
\begin{equation}
    \max_{\mathcal{P}_\textrm{pre} \subset \mathcal{F}_\textrm{pre}: |\mathcal{P}_\textrm{pre}| = n} J(\mathcal{P}_\textrm{pre}) \;.
     \label{eq:objective}
\end{equation}
This optimization is 
intractable, but a near-optimal solution can be achieved through a greedy process since the function $J$ is normalized monotone submodular~\citep{nemhauser1978analysis} when using the RBF kernel for $k(\cdot, \cdot)$, as proved in \citet{kim2016examples}.
By maximizing the score function $J(\mathcal{P}_\textrm{pre})$, selected prototypes closely approximate the entire distribution of domain embeddings in the queue.
At the initial stage where no previous test domain exists, we use domain embeddings of augmented test images as domain prototypes.

\subsection{Domain-invariant learning}\label{sec:model_training}
TestDG updates the model through domain-invariant learning and self-training. 
For domain-invariant learning, we employ a two-step process.
In the first step, the domain information extractor and amplifier are updated to extract domain-specific information, \ie, domain embeddings, from the encoder features. 
In the second step, the encoder is updated to be domain-invariant by aligning the current domain embeddings with the previous domain prototypes (Sec.~\ref{sec:proto_selection}).
Alternating these two steps allows the encoder to gradually become domain-invariant, and the domain information extractor and amplifier to become sensitive to the remaining domain-specific information of the encoder.
In addition, following previous work~\citep{wang2022continual,gan2023decorate,liu2023vida}, we update our model by a self-training loss $\mathcal{L}_\textrm{self}$, 
a cross-entropy loss between our model's prediction $\hat{y}$ and a pseudo label $\Tilde{y}$ generated by its exponential moving average~\citep{tarvainen2017mean}:
\begin{align}
    \mathcal{L}_\textrm{self}(x) = -\sum_{c=1}^{C}\tilde{y}_c \log \hat{y}_c,
    \label{eq:ce}
\end{align}
where $C$ is the total number of classes and $\tilde{y}$ is the softmax prediction of the EMA teacher model.
For inference, we use the prediction of the EMA teacher model as the classification prediction.
Details of the two steps are described below.

\noindent\textbf{Step 1: Extracting domain embeddings.}\label{sec:step1}
To maintain the semantic information of the encoder features during domain-invariant learning, we first extract their domain-specific information in the form of embedding vectors.
To this end, we introduce the domain information extractor
that extracts domain embeddings from the output of the encoder and projects them into a 4-dimensional embedding space,
and the domain amplifier
that amplifies the domain-specific information of the encoder to facilitate the extractor.
Let $\mathcal{F}_\textrm{cur}=\{f_i\}_{i=1}^{n}$ be the domain embeddings of test input images 
from the current test domain
and $\mathcal{P}_\textrm{pre}$ the previous domain's prototypes.
The domain information extractor and amplifier learn a domain embedding space by distinguishing
$\mathcal{F}_\textrm{cur}$ and 
$\mathcal{P}_\textrm{pre}$
with a domain discrimination loss:
\begin{align}
\mathcal{L}_{\textrm{dis}}(\mathcal{F}_\textrm{cur},\mathcal{P}_\textrm{pre}) = - \frac{1}{n}\bigg\{\displaystyle\sum_{f_i \in \mathcal{F}_\textrm{cur},}&\,\log(D(f_i)) 
+\sum_{p_i \in \mathcal{P}_\textrm{pre}}\log(1-D(p_i))\bigg\},
\label{eq:loss_adv_discriminating}
\end{align}
where $D(\cdot)$ indicates a binary classifier implemented as a three-layer MLP that serves as a domain discriminator.
Minimizing this loss encourages the domain extractor with the amplifier to learn a space where embeddings of different domains are separated and thus capture domain-specific information.

\noindent\textbf{Step 2: Learning domain-invariant features.}\label{sec:step2}
The second step aims to optimize the encoder to achieve domain invariance across diverse test domains alongside classification (Eq.~\ref{eq:ce}).
TestDG accomplishes this by aligning current domain embeddings and previous domain prototypes.
To be specific, we update the encoder and its classification head while freezing the domain information extractor and amplifier.
Given current domain embeddings $\mathcal{F}_\textrm{cur}$ and previous domain prototypes $\mathcal{P}_\textrm{pre}$, the domain-invariant learning loss is given by
\begin{align}
\mathcal{L}_{\textrm{inv}}(\mathcal{F}_\textrm{cur},\mathcal{P}_\textrm{pre}) = \frac{1}{n}\sum\limits_{ (f_i,p_i)  \in (\mathcal{F}_\textrm{cur},\mathcal{P}_\textrm{pre}) } \left\|f_i-p_i\right\|_1.
\label{eq:loss_dil}
\end{align}
By minimizing this loss, the encoder's features are trained to produce domain embeddings that are indistinguishable between different domains, making them domain-generalizable to future domains.
The current-domain embeddings and previous-domain prototypes are paired at random.

\subsection{Domain prototype update}\label{sec:proto_update}
Since the model used to extract the domain prototypes is constantly updated during CTTA, the domain prototypes extracted from the earlier model become outdated.
Thus, the domain prototypes should also be updated to reflect the update of the model, approximating prototypes extracted from the updated model.
Inspired by \citet{asadi2023prototype}, we aim to preserve the set similarity between the previous domain prototypes $\mathcal{P}_\textrm{pre}$ and the current domain embeddings $\mathcal{F}_\textrm{cur}$ to effectively keep the relationship between them intact.
Specifically, we measure the Chamfer distance between the two sets $\mathcal{P}_\textrm{pre}$ and $\mathcal{F}_\textrm{cur}$.
For iteration $t$, we update the domain prototypes $\mathcal{P}_\textrm{pre}^{t-1}$ to $\mathcal{P}_\textrm{pre}^{t}$ to ensure that the Chamfer distance between previous domain prototypes and current domain embeddings remains consistent before and after the model update.
Specifically, the domain prototypes $\mathcal{P}_\textrm{pre}^{t}$ are updated by minimizing the prototype updating loss:
\begin{align}
\mathcal{L}_\textrm{update}=|d_\textrm{CD}(\mathcal{F}_\textrm{cur}^{t-1},\mathcal{P}_\textrm{pre}^{t-1})-d_\textrm{CD}(\mathcal{F}_\textrm{cur}^{t},\mathcal{P}_\textrm{pre}^{t})|,
\label{eq:proto_update_chamfer}
\end{align}
where the Chamfer distance $d_\textrm{CD}$ is computed by
\begin{align}
d_\textrm{CD}(\mathcal{F}_\textrm{cur},\mathcal{P}_\textrm{pre}) = \frac{1}{|\mathcal{F}_\textrm{cur}|}\sum_{f_i \in \mathcal{F}_\textrm{cur}}\min_{p_j \in \mathcal{P}_\textrm{pre}}\left\|p_j-f_i\right\|_2 \nonumber\\
     + \frac{1}{|\mathcal{P}_\textrm{pre}|}\sum_{p_j \in \mathcal{P}_\textrm{pre}}\min_{f_i \in \mathcal{F}_\textrm{cur}}\left\|p_j-f_i\right\|_2.
\label{eq:proto_update}
\end{align}
$\mathcal{P}_\textrm{pre}^{t}$ is initialized randomly at each iteration and optimized for a single step on $\mathcal{L}_\textrm{update}$.

\begin{figure*}[t]
\centering
\includegraphics[width=1.0\linewidth]{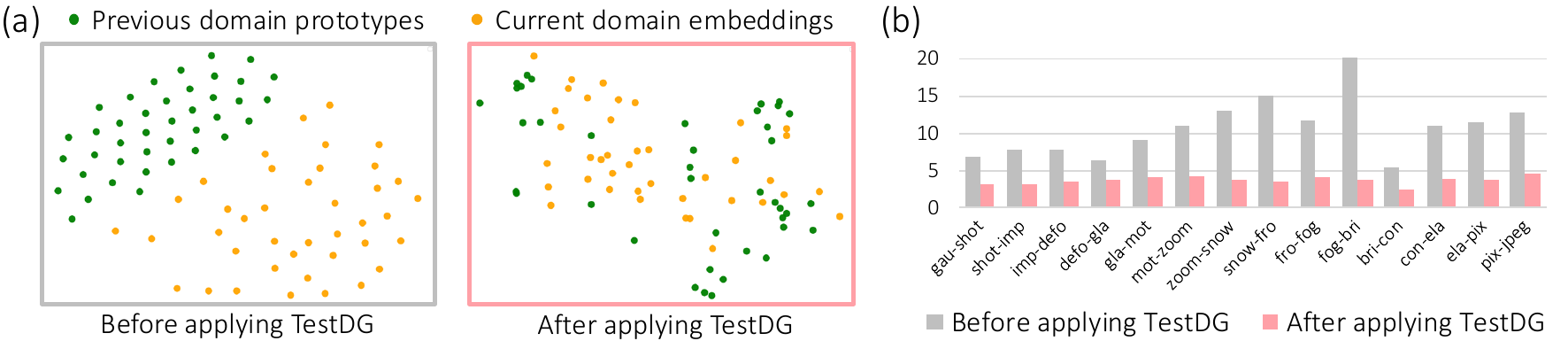}
\caption{Analysis on the impact of domain-invariant learning in TestDG. (a) 2D visualization of the distribution of domain embeddings from different domains before and after TestDG. (b) The domain gap between different domains measured by Chamfer distance before and after applying TestDG.}
\label{fig:analysis}
\end{figure*}

\begin{figure*}[t]
    \centering
    \includegraphics[width=1.0\linewidth]{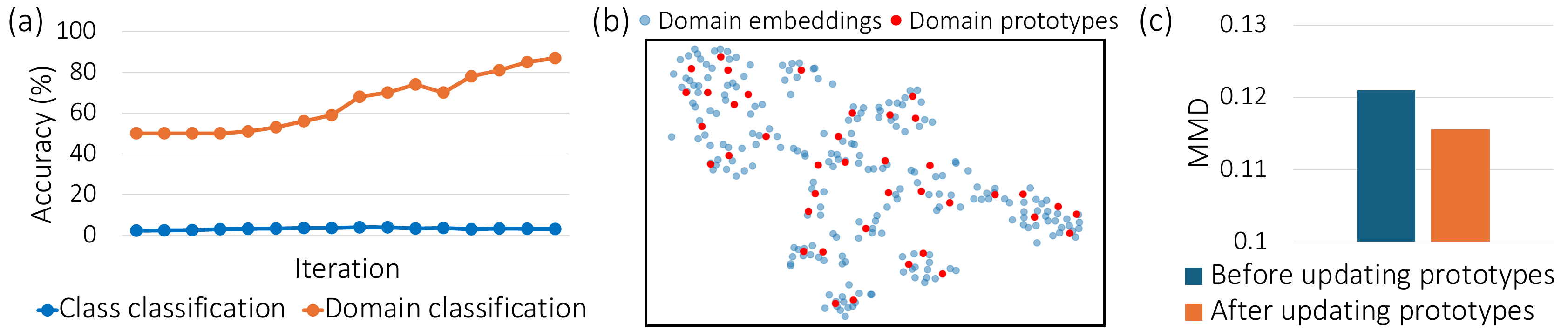}
    \caption{Analysis on domain embeddings and prototypes. (a) Analysis of class classification and domain classification accuracy using domain embeddings during CTTA iterations. (b) 2D visualization of domain embeddings and domain prototypes. 
    (c) MMD between the target and original prototypes before the update and that between the target and updated prototypes.
    }
    \label{fig:analysis_domainembedding}
\end{figure*}
\subsection{Empirical justification}

\noindent\textbf{Domain-invariant learning.} We analyze the impact of domain-invariant learning both qualitatively and quantitatively.
\Fig{analysis}(a) provides a $t$-SNE visualization~\citep{MaatenNov2008} of domain embeddings from different domains (\ie, Gaussian and Shot noise) before and after applying TestDG, demonstrating that TestDG effectively reduces the gap between different domains.
\Fig{analysis}(b) shows the Chamfer distance between domain embeddings from different domains before and after applying TestDG, indicating that the distance decreases as intended after the domain-invariant learning.

\noindent\textbf{Domain embedding.} 
In \Fig{analysis_domainembedding}(a), we empirically verify that domain embeddings capture domain-specific information more prominently than content information. 
To assess the content information within the domain embeddings, we performed linear probing by training a linear classification layer using the domain embeddings as input
while keeping the model frozen.
The results show that domain classification using the domain embeddings consistently achieves higher accuracy than class classification, indicating that the domain embeddings predominantly encode domain-specific information with minimal content information, as intended.

\noindent\textbf{Domain prototype selection.} We qualitatively investigate how well domain prototypes represent the entire set of domain embeddings.
\Fig{analysis_domainembedding}(b) provides a $t$-SNE visualization of domain embeddings and selected domain prototypes
on the Gaussian noise domain,
showing
that our domain prototypes are 
diverse and well represent the overall distribution of the domain embeddings.

\noindent\textbf{Domain prototype update.} 
\Fig{analysis_domainembedding}(c) demonstrates the effectiveness of the prototype updating strategy. 
By forwarding the corresponding images of each prototype through the updated model, we obtain 
ideal target prototypes 
in the updated feature space.
The MMD value between these target prototypes and our updated prototypes is lower than those between target prototypes and original prototypes, where MMD values are averaged across all CTTA steps.
This result validates that our updating method effectively moves the prototypes toward their targeted representations.

\section{Experiments}
\subsection{Experimental setting}
\noindent\textbf{Evaluation benchmarks.} 
Our method was evaluated on both classification and semantic segmentation tasks.
For classification, we utilized three benchmarks~\citep{Hendrycks2019_ImageNetC}: CIFAR10-C, CIFAR100-C, and ImageNet-C, each consisting of 15 corruption types and five severity levels. 
We reported the results for the most severe level (level 5). 
For segmentation, we used Cityscapes~\citep{cityscapes} as the source and ACDC~\citep{acdc} as a test dataset, including four unseen test domains: Fog, Night, Rain, and Snow.

\noindent\textbf{Evaluation scenario.} 
Following \citet{wang2022continual}, TestDG was evaluated under an online CTTA setting where the prediction is assessed immediately after data are streamed, and the pre-trained model is adapted on the fly.
For the segmentation task, test domains are repeated cyclically for three rounds. 
In line with recent CTTA studies~\citep{liu2023vida,liu2024continual,yang2024versatile}, \textit{we reported results for all methods using a common ViT backbone}. 
This follows the standardization introduced by ViDA~\citep{liu2023vida}, which established the use of ViTs in CTTA and re-implemented previous CNN-based methods within this ViT-based framework.
For additional experiments beyond those reported by previous studies, we evaluated methods with publicly available code.
The compared methods also include pseudo-labeling (Pseudo)~\citep{lee2013pseudo}, DePT~\citep{gao2022visual}, and the method of \citet{zhu2025reshaping} (Zhu~\etal).

\noindent\textbf{Implementation details.}
We utilized ViT-base~\citep{dosovitskiy2020image} for classification tasks and Segformer-B5~\citep{xie2021segformer} for the segmentation task as the backbone.
Input images were resized to $384 \times 384$ for CIFAR10-C and CIFAR100-C, $224 \times 224$ for ImageNet-C, and to $ 960 \times 540$ for ACDC. 
The size of the queue storing the domain embeddings and the number of domain prototypes were set to 256 and 40 for CIFAR10-C, and 512 and 50 for CIFAR100-C and ImageNet-C, respectively; the number of domain prototypes equals the batch size.
The domain information extractor $\psi_\textrm{domain}$ and domain discriminator $D$ were both implemented as MLP whose activation functions are GELU~\citep{hendrycks2016gaussian} and ReLU~\citep{agarap2018deep}, respectively; $\psi_\textrm{domain}$ has two layers ($C \rightarrow 2C \rightarrow 4$) and $D$ has three layers ($4 \rightarrow 16 \rightarrow 8 \rightarrow 1$).
The domain amplifier was implemented as an adapter that includes a down-projection layer followed by an up-projection layer, with an intermediate dimension of 128, attached to the attention layers of the encoder. 
We used Adam with learning rates of $1\times10^{-4}$ and $3\times10^{-4}$ for the encoder and $6\times10^{-5}$ and $9\times10^{-6}$ for the domain amplifier on CIFAR10-C and CIFAR100-C, respectively, and SGD (momentum 0.9) with learning rates of $5\times10^{-7}$ and $2\times10^{-7}$ on ImageNet-C.
The domain information extractor and discriminator were optimized with Adam using a learning rate of $1\times10^{-5}$ for CIFAR10-C and CIFAR100-C and $2\times10^{-5}$ for ImageNet-C.
The weights of $\mathcal{L}_\textrm{dis}$ and $\mathcal{L}_\textrm{inv}$ were set to (0.05, 0.07), (0.005, 0.005), and (0.06, 0.006) for CIFAR10-C, CIFAR100-C, and ImageNet-C, respectively.
The EMA momentum of the teacher model was 0.999, except for the domain amplifier on ImageNet-C, which used 0.8.
Following \citet{wang2022continual}, we stochastically restore the adapted parameters to the source weights with a probability of 0.1 at each iteration.
The domain change detection threshold was set to 0.1, 0.16, and 0.2 for CIFAR10-C, CIFAR100-C, and ImageNet-C, respectively, and 0.1 for ACDC.
For ACDC, we used a queue size of 5 and a single domain prototype due to the memory cost of high-resolution dense features.


\begin{table*}[t]
\centering
\caption{\label{tab:cifar10}
Classification error rates~(\%) on CIFAR10-to-CIFAR10-C online CTTA task.
Gain (\%) denotes the improvement in accuracy over the source model, measured in percentage points.
}
\begin{adjustbox}{width=1\linewidth,center=\linewidth}
\begin{tabular}{@{}l|ccccccccccccccc|cc@{}}
\toprule
 Method  &
 \rotatebox[origin=c]{70}{gaussian} & \rotatebox[origin=c]{70}{shot} & \rotatebox[origin=c]{70}{impulse} & \rotatebox[origin=c]{70}{defocus} & \rotatebox[origin=c]{70}{glass} & \rotatebox[origin=c]{70}{motion} & \rotatebox[origin=c]{70}{zoom} & \rotatebox[origin=c]{70}{snow} & \rotatebox[origin=c]{70}{frost} & \rotatebox[origin=c]{70}{fog}  & \rotatebox[origin=c]{70}{brightness} & \rotatebox[origin=c]{70}{contrast} & \rotatebox[origin=c]{70}{elastic\_trans} & \rotatebox[origin=c]{70}{pixelate} & \rotatebox[origin=c]{70}{jpeg}
& Mean$\downarrow$ & Gain\\
\midrule
Source&60.1&53.2&38.3&19.9&35.5&22.6&18.6&12.1&12.7&22.8&5.3&49.7&23.6&24.7&23.1&28.2&/\\
Pseudo&59.8&52.5&37.2&19.8&35.2&21.8&17.6&11.6&12.3&20.7&5.0&41.7&21.5&25.2&22.1&26.9&+1.3\\
TENT&57.7&56.3&29.4&16.2&35.3&16.2&12.4&11.0&11.6&14.9&4.7&22.5&15.9&29.1&19.5&23.5&+4.7\\
CoTTA&58.7&51.3&33.0&20.1&34.8&20.0&15.2&11.1&11.3&18.5&4.0&34.7&18.8&19.0&17.9&24.6&+3.6\\
VDP&57.5&49.5&31.7&21.3&35.1&19.6&15.1&10.8&10.3&18.1&4.0&27.5&18.4&22.5&19.9&24.1&+4.1\\
ViDA & 52.9&  47.9 &  19.4&  11.4&  31.3&  13.3&  7.6&  7.6&  9.9&  12.5&  3.8&  26.3&  14.4&  33.9&  18.2&  20.7& +7.5\\
Continual-MAE & 30.6 & 18.9 & 11.5 & 10.4 & 22.5 & 13.9 & 9.8 & 6.6 & 6.5 & 8.8 & 4.0 & 8.5 & 12.7 & 9.2 & 14.4 & 12.6 &+15.6 \\
Versatile & 16.3 & 11.1& 9.6& 8.4 &14.6& 8.6& 5.5 &6.3 &5.7 &7.1 &3.3 &5.4 &10.9 &7.7 &12.8 &8.9& +19.3 \\
\textbf{TestDG} & 17.4 & 10.9 & 5.4 & 6.3 & 16.3 & 6.9 & 3.9 & 3.7 & 3.8 & 5.9 & 2.2 & 4.7 & 7.8 & 11.2 & 9.4 & \textbf{7.7} & \textbf{+20.5} \\
\bottomrule
\end{tabular}
\end{adjustbox}
\end{table*}
\begin{table*}[t]
\resizebox{\linewidth}{!}{
\begin{minipage}[t]{0.25\linewidth}
\centering
\caption{Classification error rates~(\%) on CIFAR100-to-CIFAR100-C CTTA.}
\label{tab:cifar100}
\vspace{-2mm}
\renewcommand{\arraystretch}{1.15}
\resizebox{1.0\linewidth}{!}{
\Huge
\begin{tabular}{@{}l|cc@{}}
\toprule
Method & Mean$\downarrow$ & Gain \\
\midrule
Source        & 35.4 & / \\
Pseudo        & 33.2 & +2.2 \\
TENT          & 32.1 & +3.3 \\
CoTTA         & 34.8 & +0.6 \\
VDP           & 32.0 & +3.4 \\
ViDA          & 27.3 & +8.1 \\
Continual-MAE & 26.4 & +9.0 \\
Versatile     & 24.0 & +11.4 \\
\textbf{TestDG} & \textbf{23.3} & \textbf{+12.1} \\
\bottomrule
\end{tabular}
}
\end{minipage}
\hfill
\hspace{2mm}
\begin{minipage}[t]{0.25\linewidth}
\centering
\caption{Classification error rates~(\%) on ImageNet-to-ImageNet-C CTTA.}
\label{tab:imagenet}
\vspace{-2mm}
\renewcommand{\arraystretch}{1.15}
\resizebox{1.0\linewidth}{!}{
\Huge
\begin{tabular}{@{}l|cc@{}}
\toprule
Method & Mean$\downarrow$ & Gain \\
\midrule
Source        & 55.8 & / \\
Pseudo        & 61.2 & -5.4 \\
TENT          & 51.0 & +4.8 \\
CoTTA         & 54.8 & +1.0 \\
VDP           & 50.0 & +5.8 \\
ViDA          & 43.4 & +12.4 \\
Continual-MAE & \textbf{42.5} & \textbf{+13.3} \\
Versatile     & 42.7 & +13.1 \\
\textbf{TestDG} & \textbf{42.5} & \textbf{+13.3} \\
\bottomrule
\end{tabular}
}
\end{minipage}
\hfill
\hspace{3mm}
\begin{minipage}[t]{0.58\linewidth}
\centering
\caption{Generalization error rates (\%) on held-out corruption domains of CIFAR10-C, CIFAR100-C, and ImageNet-C.}
\label{tab:imagenet_generalization}
\vspace{-2mm}
\Huge
\resizebox{0.8\linewidth}{!}{
\begin{tabular}{l|cc|cc|cc}
\toprule
\multirow{2}{*}{Method}
& \multicolumn{2}{c|}{CIFAR10-C}
& \multicolumn{2}{c|}{CIFAR100-C}
& \multicolumn{2}{c}{ImageNet-C} \\
\cmidrule(lr){2-3}
\cmidrule(lr){4-5}
\cmidrule(lr){6-7}
& Mean$\downarrow$ & Gain
& Mean$\downarrow$ & Gain
& Mean$\downarrow$ & Gain \\
\midrule
Source
& 25.3 & /
& 45.2 & /
& 49.9 & / \\
T3A
& 13.6 & +11.7
& 34.8 & +10.4
& 42.2 & +7.7 \\
ViDA
& 21.5 & +3.8
& 49.5 & -4.3
& 40.6 & +9.3 \\
\textbf{TestDG}
& \textbf{11.8} & \textbf{+13.5}
& \textbf{33.1} & \textbf{+12.1}
& \textbf{39.9} & \textbf{+10.0} \\
\bottomrule
\end{tabular}
}

\begin{center}
\vspace{1mm}
\caption{CTTA results under gradual domain shifts on CIFAR10-to-CIFAR10-C, evaluated by error rates (\%).}
\vspace{-2mm}
\resizebox{0.99\linewidth}{!}{
\huge
\begin{tabular}{C{2.5cm}C{4cm}C{4cm}C{6cm}C{4cm}}
\toprule
& Source & ViDA & Continual-MAE & \textbf{TestDG} \\
\midrule
Mean$\downarrow$ & 13.7 & 7.7 & 6.5 & \textbf{3.6} \\
Gain & / & +6.0 & +7.2 & \textbf{+10.1} \\
\bottomrule
\end{tabular}
}
\label{tab:gradual_shift}
\end{center}
\end{minipage}
}
\end{table*}

\subsection{Evaluation on classification tasks}

To evaluate the performance of TestDG, we conducted extensive experiments on three conventional benchmarks: CIFAR10-C, CIFAR100-C, and ImageNet-C.

\noindent\textbf{Performance on corruption benchmarks.}
Table~\ref{tab:cifar10} and Table~\ref{tab:cifar100} present the results on CIFAR10-C and CIFAR100-C.
TestDG significantly outperformed all other methods across various corruption types.
Specifically, it achieved the lowest mean error rate of 7.7\% on CIFAR10-C and 23.3\% on CIFAR100-C, showing improvements of 20.5\%p and 12.1\%p over the source model, respectively.
As shown in Table~\ref{tab:imagenet}, TestDG performed on par with the best-performing baseline, Continual-MAE, with both methods achieving a mean error of 42.5\% on ImageNet-C.

\begin{table*}[t]
\centering
\caption{\label{tab:ACDC} Segmentation mIoU results on Cityscape\sohyun{s}-to-ACDC online CTTA task.}\label{tab:CTTA_seg}
\renewcommand{\arraystretch}{1.0}
\setlength\tabcolsep{2pt}
\begin{adjustbox}{width=1\linewidth,center=\linewidth}
\begin{tabular}{l|ccccc|ccccc|ccccc|c}
\toprule

\multicolumn{1}{l|}{Time}   & \multicolumn{15}{c}{$t$ {\rightarrowfill} }                                                                              \\ \midrule
\multicolumn{1}{l|}{Round}           & \multicolumn{5}{c|}{1}    & \multicolumn{5}{c|}{2}     & \multicolumn{5}{c|}{3}  & \multirow{2}{*}{Mean$\uparrow$}    \\ \cmidrule(lr){1-16} 
Method & Fog & Night & Rain & Snow & Mean$\uparrow$ & Fog & Night & Rain & Snow  & Mean$\uparrow$ & Fog & Night & Rain & Snow & Mean$\uparrow$  \\ \midrule 
Source&69.1&40.3&59.7&57.8&56.7&69.1&40.3&59.7&  57.8&56.7&69.1&40.3&59.7& 57.8&56.7&56.7\\
TENT&69.0&40.2&60.1&57.3&56.7&68.3&39.0&60.1&     56.3&55.9&67.5&37.8&59.6&55.0&55.0&55.9\\
CoTTA &70.9&41.2&62.4&59.7&58.6&70.9&41.1&62.6&   59.7&58.6&70.9&41.0&62.7&59.7&58.6&58.6\\
DePT&71.0&40.8&58.2&56.8&56.7&68.2&40.0&55.4&53.7& 54.3&66.4&38.0&47.3&47.2&49.7&53.6\\
ECoTTA&68.5 &35.8 &62.1 &57.4 &56.0 &68.3 &35.5 &62.3 &57.4& 55.9 & 68.1 & 35.3 & 62.3 & 57.3 & 55.8 & 55.9 \\
VDP&70.5&41.1&62.1&59.5&  58.3    &70.4&41.1&62.2&59.4& 58.2     & 70.4&41.0&62.2&59.4& 58.2   &  58.2 \\
ViDA&71.6& 43.2& 66.0& 63.4& 61.1&
  73.2& 44.5& 67.0& 63.9& 62.2 & 73.2& 44.6& 67.2& 64.2& \textbf{62.3} & 61.9\\
BECoTTA &72.3 &42.0 &63.5 &60.1 &59.5& 72.4& 41.9& 63.5 &60.2 &59.5&72.3& 41.9& 63.6 &60.2&59.5 &59.5 \\
SVDP&72.1 &44.0 &65.2 &63.0 &61.1 &72.2 &44.5& 65.9 &63.5& 61.5& 72.1 &44.2 &65.6 &63.6 &61.4 &61.3\\
Continual-MAE &71.9 &44.6 &67.4& 63.2 &61.8 &71.7 &44.9 &66.5 &63.1 &61.6& 72.3& 45.4& 67.1& 63.1& 62.0& 61.8\\
Zhu~\etal &71.2 &42.3& 64.9 &62.0 &60.1 & 72.6 & 43.2 & 66.3 & 63.2 & 61.3 & 72.8 & 43.8 & 66.5 & 63.2 & 61.6 & 61.0 \\
 \textbf{TestDG} &71.9& 45.5& 66.4& 63.9& \textbf{62.0$_{\pm{0.1}}$}&
  73.2& 45.8& 67.4& 64.1& \textbf{62.6$_{\pm{0.1}}$} & 72.8& 44.7& 67.9& 63.7& \textbf{62.3$_{\pm{0.1}}$} & \textbf{62.3$_{\pm{0.1}}$}\\
 \bottomrule
\end{tabular}
\end{adjustbox}
\end{table*}

\noindent\textbf{Generalization on unseen domains.}
We evaluated generalization to held-out corruption domains on CIFAR10-C, CIFAR100-C, and ImageNet-C.
We additionally compared with T3A~\citep{iwasawa2021test}, which improves domain generalization through adaptation to unlabeled test data.
Unlike this classifier-level adjustment, TestDG updates its representations through domain-invariant learning using current-domain embeddings and previous-domain prototypes.
As shown in Table~\ref{tab:imagenet_generalization}, TestDG achieved the lowest mean error on all three benchmarks among the compared methods.
These results support the benefit of domain-invariant learning for generalization to unseen test domains.

\noindent\textbf{Gradually changing results.}
Table~\ref{tab:gradual_shift} shows results of continual test-time adaptation on a CIFAR10-C gradually changing setup~\citep{wang2022continual}, where the corruption types changed gradually over time.
In this more realistic scenario where domain boundaries are blurred, TestDG still demonstrated superior performance, notably outperforming existing methods with a mean error rate of 3.6\%.
This result highlights the effectiveness of domain-invariant learning even under gradual domain transitions.

\subsection{Evaluation on segmentation tasks}
\noindent\textbf{Performance on the adverse weather benchmark.} Table~\ref{tab:CTTA_seg} presents the segmentation results from adapting Cityscapes to ACDC, a real-world adverse condition dataset consisting of four challenging domains: Fog, Night, Rain, and Snow.
TestDG achieved the highest mean mIoU in the first two rounds and over all three rounds, and tied with ViDA for the highest in the third round.
In the first round, our method reached an average mIoU of 62.0\%, showing a 5.3\%p gain over the source model.
In subsequent rounds, TestDG maintained strong performance when revisiting previously encountered domains.
These results suggest that TestDG successfully mitigates domain shifts in semantic segmentation tasks, maintaining robustness across diverse and real-world adverse conditions.

\noindent\textbf{Generalization to unseen weather domains.}
We additionally evaluated generalization on ACDC using a leave-one-domain-out protocol.
As shown in Table~\ref{tab:acdc_dg}, TestDG achieved the highest mean mIoU among the compared methods on the held-out weather domains.
This result extends the evidence for generalization beyond synthetic image corruptions to real-world adverse weather conditions.

\subsection{In-depth analysis}


\textbf{Impact of domain amplifier.}
We investigated the impact of the domain amplifier by evaluating TestDG with and without it.
As shown in Table~\ref{tab:ablation_loss}, removing the domain amplifier significantly degrades performance, with the error rate increasing from 7.7\% to 21.9\%.
This suggests that the domain amplifier plays an important role, as it amplifies domain-specific information from each layer of the encoder, thereby enabling the encoder to focus on learning domain-invariant features.

\begin{table*}[t]

\begin{minipage}[t]{0.25\textwidth}
  \vspace{0pt}
  \centering
  \small
  \captionof{table}{Generalization performance on ACDC.}
  \label{tab:acdc_dg}
  \renewcommand{\arraystretch}{1.1}
  \resizebox{\linewidth}{!}{
    \begin{tabular}{l|c}
      \toprule
      Method & Mean mIoU $\uparrow$ \\
      \midrule
      Source & 58.3 \\
      CoTTA  & 58.5 \\
      ViDA   & 58.4 \\
      \textbf{TestDG} & \textbf{59.0} \\
      \bottomrule
    \end{tabular}
  }
\end{minipage}
\hfill
\begin{minipage}[t]{0.36\textwidth}
  \vspace{0pt}
  \centering
  \captionof{table}{Ablation study on each component of TestDG.}
  \label{tab:ablation_loss}
  \vspace{-1.7mm}
  \renewcommand{\arraystretch}{0.9}
  \resizebox{0.98\linewidth}{!}{
    \begin{tabular}{@{}ccccc|c@{}}
      \toprule
      \multicolumn{5}{c|}{$\psi_\textrm{amplifier}$}
      & Mean$\downarrow$ \\
      \midrule
      \multicolumn{5}{c|}{} & 21.9 \\
      \multicolumn{5}{c|}{\checkmark} & \textbf{7.7} \\
      \midrule
      $\mathcal{L}_\textrm{self}$
      & $\mathcal{L}_\textrm{inv}$
      & $\mathcal{L}_\textrm{dis}$
      & $J(\mathcal{P}_\textrm{pre})$
      & $\mathcal{L}_\textrm{update}$
      & Mean$\downarrow$ \\
      \midrule
      & & & & & 28.2 \\
      \checkmark & & & & & 18.0 \\
      \checkmark & \checkmark & & & & 10.5 \\
      \checkmark & \checkmark & \checkmark & & & 8.3 \\
      \checkmark & \checkmark & \checkmark
      & \checkmark & & 8.0 \\
      \checkmark & \checkmark & \checkmark
      & \checkmark & \checkmark & \textbf{7.7} \\
      \bottomrule
    \end{tabular}
  }
\end{minipage}
\hfill
\begin{minipage}[t]{0.36\textwidth}
  \vspace{0pt}
  \centering
  \includegraphics[width=\linewidth]{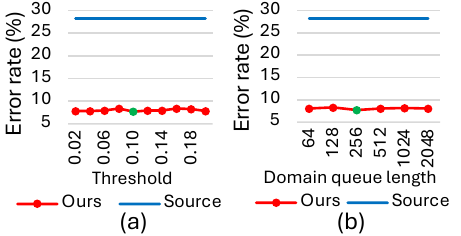}
  \vspace{-5.3mm}
  \captionof{figure}{
    Sensitivity to hyperparameters.
    (a) Threshold for domain change detection.
    (b) Queue length.
  }
  \label{fig:ablation_graph}
\end{minipage}

\end{table*}

\noindent\textbf{Ablation study of each loss.}
We conducted an ablation study to evaluate the contributions of each loss component in our method.
The results, presented in \Tbl{ablation_loss}, demonstrate that each loss contributes to the overall performance improvement. 
Notably, the domain-invariant learning loss $\mathcal{L}_\textrm{inv}$ was crucial for achieving high performance by guiding the model to learn domain-invariant features, which are essential for effective adaptation across different domains.
Additionally, the domain discrimination loss $\mathcal{L}_\textrm{dis}$, which facilitated the learning of domain embeddings, was also crucial for TestDG.
Furthermore, both the score function for domain prototype selection $J(\mathcal{P}_\textrm{pre})$ and the prototype updating loss  $\mathcal{L}_\textrm{update}$ contributed to the performance of TestDG.

\noindent\textbf{Sensitivity to the threshold for domain change detection.}
To investigate the effect of the threshold used for domain change detection, we conducted experiments varying the threshold values.
\Fig{ablation_graph}(a) summarizes the results for thresholds ranging from 0.02 to 0.20.
The performance remained stable across this range of thresholds, consistently outperforming the source model.
These results suggest that TestDG is insensitive to threshold settings, which is practically important since the optimal threshold may vary across different deployment scenarios.

\noindent\textbf{Sensitivity to the queue length $|\mathcal{F}_{\text{pre}}|$.}
We evaluated the impact of $|\mathcal{F}_{\text{pre}}|$, which specifies the length of the queue that stores previous domain embeddings.
It is designed to retain a sufficient number of embeddings to capture the distribution of the previous domain while minimizing the inclusion of excessively outdated information.
\Fig{ablation_graph}(b) presents the error rates on CIFAR10-C for six different queue lengths ranging from 64 to 2,048.
The results indicate that the error rates remain consistent across all tested values of $|\mathcal{F}_{\text{pre}}|$, demonstrating that TestDG is insensitive to the length of the queue and can maintain strong performance even with a compact memory footprint.

\begin{table*}[t]
\centering

\begin{minipage}[t]{0.30\textwidth}
  \vspace{0pt}
  \centering
  \includegraphics[width=0.92\linewidth]{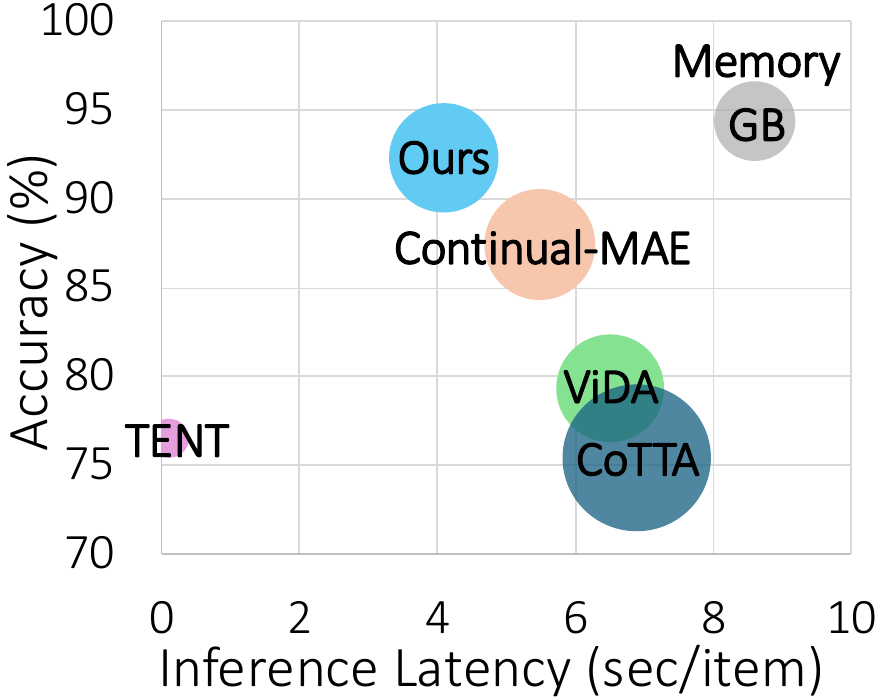}
  \captionof{figure}{Comparison of computational cost on CIFAR10-C.}
  \label{fig:efficiency}
\end{minipage}
\hfill
\begin{minipage}[t]{0.27\textwidth}
  \vspace{0pt}
  \centering
  \includegraphics[width=\linewidth]{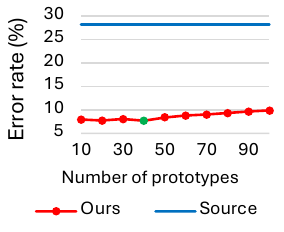}
  \captionof{figure}{Ablation on the number of domain prototypes.}
  \label{fig:ablation_proto}
\end{minipage}
\hfill
\begin{minipage}[t]{0.33\textwidth}
  \vspace{0pt}
  \centering
  \small
  \setlength{\tabcolsep}{4pt}
  \renewcommand{\arraystretch}{1.0}
  \captionsetup{skip=3pt}

  \captionof{table}{Extended ablation on thresholds.}
  \label{tab:ablation_threshold}
  \begin{subtable}[t]{\linewidth}
    \centering
    \begin{tabular*}{\linewidth}{@{\extracolsep{\fill}}lccc@{}}
      \toprule
      Length & 8192 & 16384 & 32768 \\
      \midrule
      Error & 8.1 & 8.0 & 8.1 \\
      \bottomrule
    \end{tabular*}
    \vspace{0.1mm}
    \caption{Queue length}
  \end{subtable}
  \par\vspace{1mm}
  \begin{subtable}[t]{\linewidth}
    \centering
    \begin{tabular*}{\linewidth}{@{\extracolsep{\fill}}lccc@{}}
      \toprule
      Conf & 0.4 & 0.6 & 0.8 \\
      \midrule
      Error & 8.1 & 8.3 & 8.1 \\
      \bottomrule
    \end{tabular*}
    \vspace{0.1mm}
    \caption{Confidence threshold}
  \end{subtable}
\end{minipage}
\vspace{-2mm}
\end{table*}

\noindent\textbf{Comparison of computational cost.}
We evaluated the computational cost of TestDG in terms of inference latency and memory consumption alongside prediction accuracy.
As shown in \Fig{efficiency}, while TENT~\citep{Tent} exhibited the highest computational efficiency, it had lower prediction accuracy compared to most evaluated methods.
In contrast, TestDG achieved reduced inference latency and comparable memory consumption compared to other CTTA methods (excluding TENT) while significantly outperforming them in prediction accuracy.
This indicates a favorable trade-off between computational cost and performance.

\noindent\textbf{Number of domain prototypes.}
We conduct an ablation study on the number of domain prototypes to analyze its impact on TestDG's performance.
As shown in \Fig{ablation_proto}, we vary the number of domain prototypes from 10 to 100 while keeping other hyperparameters fixed. The error rate remains nearly unchanged from 10 to 40 prototypes and increases gradually beyond 40, as using many prototypes not only increases computational overhead but also risks including outliers that may degrade the quality of domain-invariant learning. Since the variation stays within a few percentage points across the tested range, TestDG does not require careful tuning of this hyperparameter.

\noindent\textbf{Extended ablation on thresholds.}
We expand the ablation studies to include a wider range of queue lengths and domain change detection thresholds beyond those reported in Figure~\ref{fig:ablation_graph}.
Table~\ref{tab:ablation_threshold} shows that TestDG remains robust to a very wide range of hyperparameter settings, including queue lengths up to 32,768 and confidence thresholds up to 0.8, demonstrating the practical reliability of our method.

\noindent\textbf{Distance metrics for prototype update.}
We investigate the effectiveness of different distance metrics (\ie, MMD, Hausdorff, and Chamfer) for updating the prototypes in the domain embedding space.
As shown in Table~\ref{tab:distance_metrics}, Chamfer distance results in the lowest error among the three metrics.
This result supports its use in our prototype update objective.

\noindent\textbf{Number of retained previous domains.}
We investigate the effect of retaining prototypes from multiple previous domains.
As shown in Table~\ref{tab:previous_domains}, using more previous domains worsens both CTTA and held-out-domain generalization.
Since the feature space continually evolves during adaptation, prototypes from earlier domains become increasingly stale and may no longer faithfully represent their original distributions.
Consequently, aligning to these outdated references can be detrimental.
Retaining only the most recent previous domain achieves the best performance while keeping memory usage constant regardless of the stream length.

\noindent\textbf{Comparison of prototype selection strategies.}
We compared our prototype selection strategy with FIFO, random selection, and reservoir sampling under the same memory budget.
As shown in Table~\ref{tab:buffer_selection}, our method achieved both the lowest CTTA error and the smallest squared MMD between the selected prototypes and the previous-domain embeddings.
These results show that our selection better approximates the previous-domain embedding distribution while improving adaptation.


\noindent\textbf{Comparison of learnable parameters.}
We compare the number of learnable parameters between TestDG and other CTTA methods in Table~\ref{tab:param_compare}.
TestDG requires 7.0M more learnable parameters than CoTTA and Continual-MAE, and slightly fewer than ViDA, while TENT updates only the normalization parameters.
Despite having more parameters than some baselines, TestDG achieves substantially lower error, indicating that its gains do not stem from parameter count alone.


\begin{table*}[t]
\begin{minipage}[t]{0.37\textwidth}
    \centering
    \captionof{table}{Distance for prototype update.}
    \vspace{-3mm}
    \label{tab:distance_metrics}
    \scalebox{0.8}{
    \resizebox{0.97\linewidth}{!}{
    \begin{tabular}{lccc}
        \toprule
        Metric & MMD & Hausdorff & Chamfer \\
        \midrule
        Error & 8.2 & 8.1 & \textbf{7.7} \\
        \bottomrule
    \end{tabular}}}
    \vspace{-1mm}
    \captionof{table}{Effect of previous domains.}
    \label{tab:previous_domains}
    \scalebox{0.8}{
    \resizebox{0.97\linewidth}{!}{
    \begin{tabular}{lcc}
        \toprule
        Previous domains & CTTA $\downarrow$ & DG $\downarrow$ \\
        \midrule
        1 (ours) & \textbf{7.7} & \textbf{11.8} \\
        3        & 8.3 & 12.7 \\
        5        & 8.2 & 13.1 \\
        \bottomrule
    \end{tabular}}}
\end{minipage}
\hfill
\begin{minipage}[t]{0.31\textwidth}
    \centering
    \captionof{table}{Comparison of prototype selection strategies under the same memory budget.}
    \label{tab:buffer_selection}
  \renewcommand{\arraystretch}{1.11}
    \resizebox{\linewidth}{!}{
    \begin{tabular}{lcc}
        \toprule
        Selection & Mean error (\%) $\downarrow$ & MMD$^2$ $\downarrow$ \\
        \midrule
        FIFO      & 8.8 & 0.0256 \\
        Random    & 8.6 & 0.0248 \\
        Reservoir & 8.4 & 0.0251 \\
        Ours      & \textbf{7.7} & \textbf{0.0216} \\
        \bottomrule
    \end{tabular}}
\end{minipage}
\hfill
\begin{minipage}[t]{0.30\textwidth}
    \centering

    \captionof{table}{Error rates and learnable parameters.}
    \label{tab:param_compare}
  \renewcommand{\arraystretch}{1.12}
    \resizebox{\linewidth}{!}{
    \begin{tabular}{lcc}
        \toprule
        Method & Error (\%) & Params \\
        \midrule
        TENT          & 23.5 & 0.038M \\
        CoTTA         & 24.6 & 86.1M \\
        ViDA          & 20.7 & 93.2M \\
        Continual-MAE & 12.6 & 86.1M \\
        \textbf{TestDG} & \textbf{7.7} & 93.1M \\
        \bottomrule
    \end{tabular}}
\end{minipage}
\end{table*}

\section{Conclusion}
We have introduced TestDG, a novel online test-time domain generalization framework for CTTA.
Unlike existing CTTA methods that focus only on adaptation to the current test domain and thus often lack generalization to arbitrary future test domains, TestDG aims to learn domain-invariant features during testing through a carefully designed model architecture and test-time training strategy.
To achieve this, we proposed a domain information extractor and domain amplifier that separate domain-specific information from encoder features, along with a domain prototype selection mechanism based on submodular optimization and a Chamfer distance-based prototype update strategy.
Through extensive experiments on four public CTTA benchmarks, including both classification (CIFAR10-C, CIFAR100-C, ImageNet-C) and semantic segmentation (Cityscapes-to-ACDC) tasks, we have shown that TestDG improves generalization to the evaluated unseen domains, ensures robust adaptation in dynamically changing test environments, and consequently achieves state-of-the-art or competitive performance.
Furthermore, comprehensive ablation studies confirm the contribution of each component.

\vspace{1mm}\noindent\textbf{Limitations and future work.} TestDG is less robust to initial corruptions due to limited exposure to diverse domain shifts at the early stages of adaptation, and we believe early adaptation strategies could mitigate this issue.
Also, TestDG currently uses only the most recent previous domain, as prototypes from older domains become outdated as the model evolves (Table~\ref{tab:previous_domains}); maintaining up-to-date information from multiple past domains with minimal memory overhead is a promising direction for future work.
In addition, since TestDG relies on self-training with pseudo-labels, severe label shift can degrade its performance; on a CIFAR10-C stream where three classes dominate the samples, its error increases by 1.2\%p compared with a class-balanced stream.

\vspace{3mm}
{\small
\noindent \textbf{Acknowledgement.}
This work was supported by 
the NRF grant (RS-2026-25618525), 
the IITP grants 
(RS-2022-II220290, 
RS-2022-II220926, 
RS-2024-00509258, RS-2024-00469482, 
RS-2024-00457882, 
RS-2026-25518317, 
RS-2019-II191906), 
the Advanced GPU Utilization Support Program (02-26-01-0372), and 
the AI Computing Support Project for R\&D (RS-2026-25505492)
funded by Ministry of Science and ICT, Korea.
}

\bibliography{cvlab_kwak}
\bibliographystyle{tmlr}

\end{document}